# Few-Shot Learning with Adaptive Weight Masking in Conditional GANs

Jiacheng Hu
Tulane University
New Orleans, USA

Zhen Qi
Northeastern University
Boston, USA

Jianjun Wei
Washington University in St. Louis
St Louis, USA

Jiajing Chen
New York University
New York, USA

Runyuan Bao
Johns Hopkins University
Baltimore, USA

Xinyu Qiu*
Northeastern University
Seattle, USA

*Abstract*—Deep learning has revolutionized various fields, yet its efficacy is hindered by overfitting and the requirement of extensive annotated data, particularly in few-shot learning scenarios where limited samples are available. This paper introduces a novel approach to few-shot learning by employing a Residual Weight Masking Conditional Generative Adversarial Network (RWM-CGAN) for data augmentation. The proposed model integrates residual units within the generator to enhance network depth and sample quality, coupled with a weight mask regularization technique in the discriminator to improve feature learning from small-sample categories. This method addresses the core issues of robustness and generalization in few-shot learning by providing a controlled and clear augmentation of the sample space. Extensive experiments demonstrate that RWM-CGAN not only expands the sample space effectively but also enriches the diversity and quality of generated samples, leading to significant improvements in detection and classification accuracy on public datasets. The paper contributes to the advancement of few-shot learning by offering a practical solution to the challenges posed by data scarcity and the need for rapid generalization to new tasks or categories.

## I. Introduction

Deep learning has achieved outstanding results in applications such as object detection [1], semantic segmentation [2], image classification [3], and natural language processing [4], with its performance even surpassing that of humans in some tasks. However, as researchers hope to achieve true artificial intelligence through deep learning, the limitations of deep learning have also begun to emerge. Deep learning models are easily prone to overfitting due to the severe lack of data in certain real-world scenarios. Moreover, training an effective deep learning model usually requires a large amount of annotated samples, which is a major challenge for many practical applications. More critically, in real-world scenarios, the distribution of training data and test data is not always consistent, and there is often a significant difference, which can have a negative impact on the model's generalization ability.

In contrast, the human neural system exhibits far superior performance in generalization capability, especially in processing small or single samples, and can quickly recognize and understand even new categories. This ability is largely attributed to the high degree of generalization of the neural system, which allows it to quickly identify and interpret new situations or information, enabling humans to easily cope with various learning tasks. To endow deep learning networks with this human-like ability of rapid learning and learning to learn, the concept of few-shot learning (FSL) [5] has emerged in recent years as a research hotspot, aiming to solve the problem of deep learning models being able to quickly learn and efficiently generalize to new tasks or new categories in the case of limited data.

The sample space augmentation method is mainly aimed at performing various transformations on the training samples, with the goal of obtaining richer training data to expand the training set, thereby effectively improving the learning ability and generalization performance of deep network models.

The new data augmentation approaches have shown excellent results in various fields, significantly improving the accuracy, robustness, and generalization ability of the models. However, in practical applications, many experimental designs require customized data augmentation methods according to the specific tasks and data characteristics. Therefore, how to efficiently generate diverse new-class data through generative models from limited samples, and thus break through the barrier of few-shot learning, is a core issue that the current new data augmentation approaches need to solve.

To address the shortcomings of the existing sample space augmentation-based few-shot learning methods, this paper proposes an improved scheme from the perspective of new data augmentation, as follows:

The core problem of few-shot learning lies in the low robustness and poor generalization of the model due to the lack of sufficient data. Although the existing generative models can generate large-scale data from a small number of samples, these generated data often lack controllability and clarity. To address this issue, this paper proposes a conditional generative adversarial network (RWM-CGAN) data augmentation method for few-shot learning, which combines residual networks and weight mask regularization techniques.

The proposed model introduces residual units in the generator, which improves the network depth of the generator and the quality of the generated samples. The discriminator adopts a weight mask regularization method, effectively suppressing the interference points in the differential images and enhancing the learning of the small-sample category features.

## II. BACKGROUND

Few-shot learning (FSL) has become an important research direction in deep learning due to its ability to enable models to generalize from limited data. Traditional deep learning models often require large datasets to achieve generalization, but FSL focuses on overcoming the data scarcity challenge. Liu et al. [6] addressed few-shot learning in product description generation by calibrating model learning, demonstrating its efficacy in novel domains. Their work highlights the need for tailored models that adapt to specific few-shot learning scenarios.

In parallel, the role of data augmentation has gained attention for enhancing model generalization in low-data regimes. Data augmentation techniques aim to enrich training data to improve the robustness of deep models, as seen in the work of Zhu et al. [7], who introduced an attention-based network for improving segmentation accuracy in medical imaging through augmented feature learning. The augmentation approaches have been critical in advancing various applications, including image classification and segmentation [8].

Conditional Generative Adversarial Networks (CGANs) have become a powerful tool for data augmentation in FSL, particularly when the goal is to generate synthetic samples that mimic the distribution of real data. The integration of CGANs into FSL has shown promising results in creating diverse, high-quality samples for robust learning [9]. In this context, our work further extends CGANs by incorporating residual networks and adaptive weight masking in the generator and discriminator, respectively, to address the shortcomings of existing methods in sample clarity and control. The residual connections improve the depth of the generator, as suggested by previous research on convolutional neural networks [10], while the weight masking technique enhances the learning of small sample categories.

Optimization techniques in deep learning are crucial for improving the efficiency of training and model performance. Zheng et al. [11] explored adaptive friction techniques to enhance deep learning optimizers, which is particularly useful for training large-scale models like GANs. Additionally, Wang et al. [12] utilized long short-term memory (LSTM) networks to overcome limitations in financial forecasting, showcasing the adaptability of optimization techniques across diverse domains.

Graph neural networks (GNNs) and hypergraph models have also been applied to enhance model learning in complex data structures, such as those seen in sequential medical predictions [13]. These models provide additional layers of abstraction, improving the overall learning and prediction accuracy. In contrast, Yao et al. [14] introduced 3D scene completion using monocular vision, showing how deep learning techniques can be adapted for specialized tasks requiring high-level scene understanding. Finally, recent advancements in neural network architectures and feature extraction methods have significantly influenced tasks like stock market prediction [15], text classification [16], and emotional analysis using large language models [17]. These studies emphasize the importance of selecting the appropriate model architecture and optimization strategy to improve model accuracy in real-world scenarios.

## III. METHOD

This paper proposes a small-sample generation model based on conditional generative adversarial networks, aiming to expand the small-sample data. On the basis of the original conditional generative adversarial network structure, the model introduces residual units in the generator to improve the network depth of the generator and the quality of the generated samples. At the same time, the discriminator combines the weight mask regularization method, which effectively suppresses the interference points in the differential images and enhances the small-sample features. By training the proposed small-sample generation model, the paper obtained synthetic samples, successfully expanding the small-sample data and improving the diversity of the generated samples.

*A. Improved Conditional Generative Adversarial Network Model Framework*

The improved model framework, as illustrated in Figure 1, is a combination of CGAN [18], Residual Unit [19], and Weight Mask [20], termed RWM-CGAN. The improvements primarily encompass three aspects:

(1) The model adopts a Convolutional Neural Network (CNN) structure.

(2) Residual units are incorporated into the generator, increasing the network depth.

(3) Weight mask techniques are employed in the discriminator.

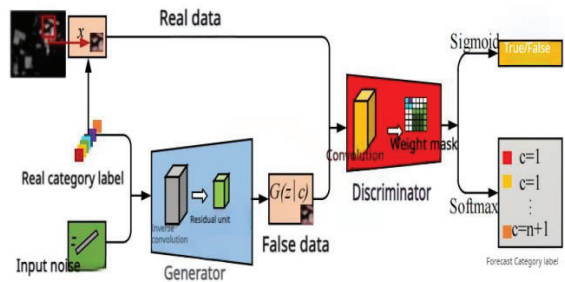

Figure 1 RWM-CGAN model framework

*B. Design Details of the Improved Conditional Generative Adversarial Network*

The original CGAN employed a fully connected structure when processing image generation tasks, resulting in poor quality of generated images. To address this issue, this paper introduces a series of improvements to the generator and discriminator network structures of CGAN, thereby enhancing the quality of generated images and the model's training efficiency.

*1) RWM-CGAN Generator Design*

RWM-CGAN incorporates residual blocks into the generator to optimize the network architecture design. The basic composition of the generator residual block is illustrated in Figure 2.

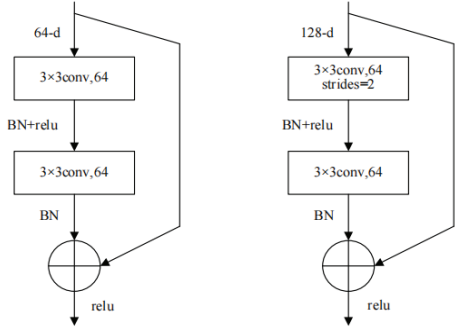

Figure 2. Generator Residual block

This approach introduces residual blocks into the generator to optimize the network architecture design. Each residual block comprises two convolutional layers with kernel sizes and strides of 3 and 1, respectively. To achieve identity mapping, ReLU activation functions and skip connections are employed between the convolutional layers. This network structure design effectively increases network depth and enhances feature extraction capabilities, thereby improving the quality of generated images. When the input and output dimensions of the residual block do not match, a convolutional operation with a kernel size of 1 and stride of 2 is used for dimension matching. This design maintains network structure consistency while avoiding performance degradation due to dimension mismatch. Furthermore, Batch Normalization (BN) layers are introduced in the residual blocks. BN layers effectively mitigate internal covariate shift issues, accelerate model convergence, and improve training efficiency.

*2) RWM-CGAN Discriminator Design*

With the continuous development of deep convolutional neural network technology, GANs have been widely applied in image processing, particularly for detecting image defects or anomalies. Ideally, compared to defect-free standard samples, each pixel value in the difference image of the GAN template should be small. However, errors may occur in practical applications, mainly because the GAN template generator may produce reconstruction errors during training due to its data-driven nature, resulting in differences between the template and actual samples. To reduce image difference interference and improve fitting accuracy, GAN model designs have become increasingly complex. However, this complexity also slows down model convergence and affects stability. To address this issue, a new solution has been proposed in the field: the weight mask regularization method. During training, this method renders some neurons ineffective with a certain probability, thereby reducing the number of network parameters. This process can be implemented by introducing a weight mask that zeroes out some weights with a certain probability during training. This method effectively reduces network complexity, improves network convergence speed and stability, and limits the range of weight values in the network, thus preventing overfitting. This approach is particularly important when detecting pixel distribution in the different images of qualified samples.

The weight mask primarily involves two parameters: $d_m(i,j,k)$ and $p^{-1}(i,j,k)$, $d_m(i,j,k)$ represents the pixel value in the difference image corresponding to the $m$-th qualified sample image in the training set, while $p^{-1}(i,j,k)$ is the number of average feature maps constructed. The weighted mask image $p^{-1}(i,j,k)$ can be represented by calculation formulas as follows:

$$p(i,j,k) = \frac{1}{M} \sum_{m=1}^{M} d_m(i,j,k)$$

$$p'(i,j,k) = \frac{p^{-1}(i,j,k) - min(p^{-1})}{max(p^{-1}) - min(p^{-1})}$$

Introducing a weight mask strategy in the discriminator is an effective way to fully utilize spatial information from small samples. This strategy can be applied to the discriminator in CGAN, which typically has a similar network structure using a fully convolutional network. To increase network stability, stridden convolutions can be introduced to replace pooling layers, and fully connected layers can be removed. The discriminator simultaneously receives fake data generated by the generator and real data as input, which then reaches the output end after convolution operations.

The process of extracting data features using weight masks is shown in Figure 3, including the following steps: First, the input sample and corresponding CGAN template undergo differential operations to highlight parts different from the template; Second, all qualifying samples in the training dataset are differentiated to further emphasize differences from the template; Then, these difference images are averaged to obtain a representative average feature map of sample differences; Finally, through inverse calculation, normalization, and other processing, a weighted mask image is obtained. In the average feature map, interfering pixel points have relatively low weighting values.

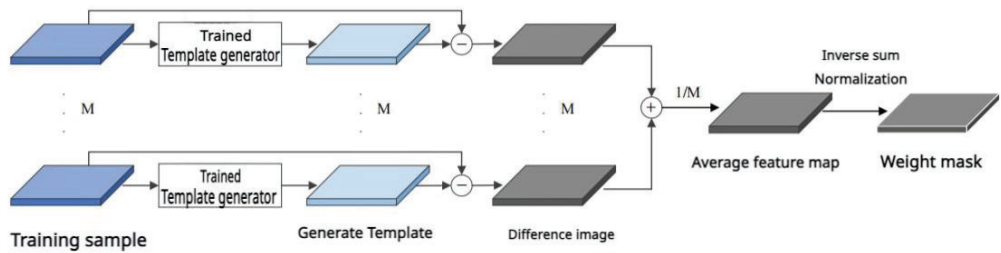

Figure 3. Flow chart of weight mask principle

## IV. EXPERIMENT

### A. Dataset

To assess the performance improvements of RWM-CGAN compared to CGAN, this study utilizes the MNIST dataset, originally employed in the CGAN paper. The MNIST dataset, compiled by the National Institute of Standards and Technology (NIST), comprises handwritten digits (0–9) contributed by 250 individuals. It includes a training set of 60,000 grayscale images, each with dimensions of 28×28 pixels, and a test set with 10,000 images. The dataset is managed using the Linked Data methodology[21], which harmonizes multiple data formats—a key factor in academic research. This structured approach enhances cross-referencing capabilities across datasets, significantly improving interoperability. Such functionality is particularly advantageous in fields like machine learning and artificial intelligence, where the quality of data plays a critical role in optimizing model training and ensuring accurate results.

### B. Experiment Results Analysis

This paper quantitatively assesses the sample clarity, diversity, similarity, and impact on the original data distribution using Inception Score (IS) and Fréchet Inception Distance (FID) [22]. The well-trained model randomly generates 2000 samples, and these samples are evaluated 10 times on average for IS, where a higher value indicates more realistic and diverse generated images. A lower FID score means that the generated data distribution is closer to the real data distribution, having a greater impact and being more similar to the original images.

The IS and FID scores for the CGAN and RWM-CGAN models are shown in Tables I and II, respectively.

TABLE I COMPARISON RESULTS OF IS BETWEEN CGAN AND RWM-CGAN

| Class | CGAN | RWM-CGAN |
|---|---|---|
| 0 | 6.481 | 6.698 |
| 1 | 6.196 | 6.768 |
| 2 | 6.366 | 6.964 |
| 3 | 6.449 | 7.037 |
| 4 | 6.209 | 6.718 |
| 5 | 6.352 | 6.944 |
| 6 | 6.565 | 6.804 |
| 7 | 6.306 | 7.369 |
| 8 | 6.646 | 7.234 |
| 9 | 6.577 | 7.126 |
| mean | 6.415 | 6.966 |

TABLE II COMPARISON RESULTS OF FID BETWEEN CGAN AND RWM-CGAN

| Class | CGAN | RWM-CGAN |
|---|---|---|
| 0 | 13.296 | 10.125 |
| 1 | 13.820 | 9.083 |
| 2 | 11.491 | 8.557 |
| 3 | 14.164 | 10.119 |
| 4 | 12.654 | 9.499 |
| 5 | 13.095 | 9.139 |
| 6 | 16.391 | 9.877 |
| 7 | 15.889 | 18.978 |
| 8 | 12.201 | 9.208 |
| 9 | 14.209 | 6.674 |
| mean | 13.721 | 10.126 |

Among the ten classes of numbers from 0 to 9, the IS scores for the images generated by the RWM-CGAN are the highest in each category. In categories "3" and "5", the IS scores of the RWM-CGAN improved by 9.12% and 9.32% compared to the CGAN model, respectively; in category "7", the IS score of the RWM-CGAN improved by 16.85% compared to the CGAN model. As can be seen from Table II, among the ten classes of numbers from 0 to 9, the FID scores of the images generated by the RWM-CGAN are the lowest in all classes except for class "7", indicating that the overall performance of the improved model is superior. Specifically, in class "9", the FID of the RWM-CGAN has decreased by 53.03%.

From the quantitative analysis above, it is evident that the RWM-CGAN data augmentation method proposed in this paper not only expands the sample set but also further enhances the ability to generate image details. However, as can be observed from the table above, for the category of the number "7", the FID value of RWM-CGAN is higher than that of CGAN. The FID measures the similarity of the samples and their impact on the original data. This situation indicates that there is still room for improvement in the way RWM-CGAN enhances the influence of the sample on the original data distribution.

TABLE III RWM-CGAN ABLATION EXPERIMENT IS AND FID RESULTS

| Method | IS | FID |
|---|---|---|
| baseline(CGAN) | 6.415 | 13.721 |
| baseline+RU | 6.504 | 12.722 |
| baseline+WM | 6.594 | 9.174 |
| baseline+RU+WM | 6.966 | 10.126 |

To demonstrate the impact of the residual units and weight masking on the experimental generation effect, an ablation study was conducted in this paper, with each set of experiments trained on the merged dataset. Table III shows the quantitative assessment of different network structures on the MNIST dataset, where the baseline model is CGAN, "RU" represents the residual units, and "WM" stands for the weight masking mechanism.

## V. CONCLUSION

This paper introduces the Residual Weight Masking Conditional Generative Adversarial Network (RWM-CGAN) as a novel and effective approach to addressing the persistent challenges of few-shot learning, particularly in scenarios where data is scarce and generalization is critical. By incorporating residual units within the generator, the model is able to enhance the network's depth, thereby producing higher quality and more diverse samples that better represent the underlying data distribution. Simultaneously, the discriminator's use of weight mask regularization serves to suppress noise and improve feature learning from small-sample categories, ensuring that the model can effectively capture and generalize from limited data. The extensive experimental results demonstrate the superiority of the RWM-CGAN approach, showing significant improvements in both sample generation and downstream tasks such as detection and classification. These findings highlight the model's ability to not only expand the sample space with greater control and clarity but also to enhance the robustness and accuracy of few-shot learning models. Ultimately, the RWM-CGAN method provides a powerful tool for overcoming the limitations of existing data augmentation strategies, offering a practical and scalable solution for advancing the field of deep learning in contexts where rapid adaptation to new tasks or categories is essential. This approach offers a novel and effective solution for overcoming the inherent limitations of current data augmentation techniques in scenarios with sparse training data, contributing to the broader goal of achieving rapid and reliable generalization in deep learning models.

REFERENCES

[1] Gui S, Song S, Qin R, et al. Remote sensing object detection in the deep learning era—a review[J]. Remote Sensing, 2024, 16(2): 327.

[2] Cheng J, Deng C, Su Y, et al. Methods and datasets on semantic segmentation for Unmanned Aerial Vehicle remote sensing images: A review[J]. ISPRS Journal of Photogrammetry and Remote Sensing, 2024, 211: 1-34.

[3] Chen K, Chen B, Liu C, et al. Rsmamba: Remote sensing image classification with state space model[J]. IEEE Geoscience and Remote Sensing Letters, 2024.

[4] Shamshiri A, Ryu K R, Park J Y. Text mining and natural language processing in construction[J]. Automation in Construction, 2024, 158: 105200.

[5] Fei-Fei L, Fergus R, Perona P. One-shot learning of object categories[J]. IEEE Transactions on Pattern Analysis and Machine Intelligence, 2006,28(4):594-611.

[6] Z. Liu, M. Wu, B. Peng, Y. Liu, Q. Peng, and C. Zou, "Calibration Learning for Few-shot Novel Product Description," in Proc. 46th Int. ACM SIGIR Conf. Res. Develop. Information Retrieval, pp. 1864-1868, July 2023.

[7] Z. Zhu, Y. Yan, R. Xu, Y. Zi, and J. Wang, "Attention-Unet: A Deep Learning Approach for Fast and Accurate Segmentation in Medical Imaging," J. Comput. Sci. Softw. Appl., vol. 2, no. 4, pp. 24-31, 2022.

[8] M. Xiao, Y. Li, X. Yan, M. Gao, and W. Wang, "Convolutional Neural Network Classification of Cancer Cytopathology Images: Taking Breast Cancer as an Example," in Proc. 2024 7th Int. Conf. Mach. Vision Appl., pp. 145-149, Mar. 2024.

[9] Y. Wei, X. Gu, Z. Feng, Z. Li, and M. Sun, "Feature Extraction and Model Optimization of Deep Learning in Stock Market Prediction," J. Comput. Technol. Softw., vol. 3, no. 4, 2024.

[10] J. Wang, S. Hong, Y. Dong, Z. Li, and J. Hu, "Predicting Stock Market Trends Using LSTM Networks: Overcoming RNN Limitations for Improved Financial Forecasting," J. Comput. Sci. Softw. Appl., vol. 4, no. 3, pp. 1-7, 2024.

[11] H. Zheng, B. Wang, M. Xiao, H. Qin, Z. Wu, and L. Tan, "Adaptive Friction in Deep Learning: Enhancing Optimizers with Sigmoid and Tanh Function," arXiv preprint arXiv:2408.11839, 2024.

[12] W. Yang, Z. Wu, Z. Zheng, B. Zhang, S. Bo, and Y. Yang, "Dynamic Hypergraph-Enhanced Prediction of Sequential Medical Visits," arXiv preprint arXiv:2408.07084, 2024.

[13] X. Yan, W. Wang, M. Xiao, Y. Li, and M. Gao, "Survival Prediction Across Diverse Cancer Types Using Neural Networks," in Proc. 2024 7th Int. Conf. Mach. Vision Appl., pp. 134-138, Mar. 2024.

[14] J. Yao, C. Li, K. Sun, Y. Cai, H. Li, W. Ouyang, and H. Li, "Ndc-scene: Boost Monocular 3d Semantic Scene Completion in Normalized Device Coordinates Space," in Proc. 2023 IEEE/CVF Int. Conf. Comput. Vision (ICCV), pp. 9421-9431, Oct. 2023.

[15] B. Wang, Y. Dong, J. Yao, H. Qin, and J. Wang, "Exploring Anomaly Detection and Risk Assessment in Financial Markets Using Deep Neural Networks," Int. J. Innov. Res. Comput. Sci. Technol., vol. 12, no. 4, pp. 92-98, 2024.

[16] E. Gao, H. Yang, D. Sun, H. Xia, Y. Ma, and Y. Zhu, "Text Classification Optimization Algorithm Based on Graph Neural Network," arXiv preprint arXiv:2408.15257, 2024.

[17] H. Yang, Y. Zi, H. Qin, H. Zheng, and Y. Hu, "Advancing Emotional Analysis with Large Language Models," J. Comput. Sci. Softw. Appl., vol. 4, no. 3, pp. 8-15, 2024.

[18] Chrysos G G, Kossaifi J, Zafeiriou S. Robust conditional generative adversarial networks[J]. arXiv preprint arXiv:1805.08657, 2018.

[19] He K, Zhang X, Ren S, et al. Deep residual learning for image recognition[C]//Proceedings of the IEEE conference on computer vision and pattern recognition. 2016: 770-778.

[20] Mallya A, Davis D, Lazebnik S. Piggyback: Adapting a single network to multiple tasks by learning to mask weights[C]//Proceedings of the European conference on computer vision (ECCV). 2018: 67-82.

[21] Li, Y., Yan, X., Xiao, M., Wang, W., & Zhang, F. (2023, December). Investigation of creating accessibility linked data based on publicly available accessibility datasets. In Proceedings of the 2023 13th International Conference on Communication and Network Security, (pp. 77-81).

[22] Barratt, S., & Sharma, R. (2018). A note on the inception score. arXiv preprint arXiv:1801.01973.